\documentclass[journal]{IEEEtran}

\ifCLASSOPTIONcompsoc
  \usepackage[nocompress]{cite}
\else
  \usepackage{cite}
\fi
\ifCLASSINFOpdf
\else
\fi
\usepackage{colortbl}
\usepackage{algorithm}
\usepackage{algorithmicx}
\usepackage{algpseudocode}
\usepackage{lipsum}
\usepackage{amsmath}
\usepackage{combelow}
\usepackage[normalem]{ulem}
\usepackage{amsfonts}
\usepackage{tabu}                      
\usepackage{booktabs}                  
\usepackage{xcolor}
\usepackage{multirow}
\usepackage{booktabs}
\usepackage{amssymb}
\usepackage{dsfont}
\usepackage{soul}
\ifCLASSINFOpdf
  \usepackage[pdftex]{graphicx}
  \DeclareGraphicsExtensions{.pdf,.jpeg,.png}
	\usepackage{epstopdf}
\else
  \usepackage[dvips]{graphicx}
  \DeclareGraphicsExtensions{.eps}
  \DeclareGraphicsExtensions{.tiff}
\fi

\definecolor{inchworm}{rgb}{0.7, 0.93, 0.36}

\usepackage{tabularx}
\usepackage{tcolorbox}
\usepackage{float}
\usepackage{amsmath}
\usepackage{array}
\newcolumntype{P}[1]{>{\centering\arraybackslash}p{#1}}
\newcolumntype{M}[1]{>{\centering\arraybackslash}m{#1}}

\begin{document}
%
\title{Automated Grading of Students' Handwritten Graphs: A Comparison of Meta-Learning and Vision-Large Language Models 
}
%
%

\author{Behnam Parsaeifard, Martin Hlosta, and Per Bergamin 
\IEEEcompsocitemizethanks{\IEEEcompsocthanksitem B. Parsaeifard, M. Hlosta, and P. Bergamin are with the Institute for Research in Open-, Distance- and eLearning, Swiss Distance University of Applied Sciences, Brig, CH-3900, Switzerland (e-mail addresses: behnam.parsaeifard@ffhs.ch, martin.hlosta@ffhs.ch, and per.bergamin@ffhs.ch). 
P. Bergamin is also affiliated with the North-West University, Potchefstroom, 2531, South Africa.\protect\\
}
}

\IEEEtitleabstractindextext{%
\begin{abstract}

With the rise of online learning, the demand for efficient and consistent assessment in mathematics has significantly increased over the past decade. 
Machine Learning (ML), particularly Natural Language Processing (NLP), has been widely used for autograding student responses, particularly those involving text and/or mathematical expressions. 
However, there has been limited research on autograding responses involving students' handwritten graphs, despite their prevalence in Science, Technology, Engineering, and Mathematics (STEM) curricula. In this study, we implement multimodal meta-learning models for autograding images containing students' handwritten graphs and text. We further compare the performance of Vision Large Language Models (VLLMs) with these specially trained meta-learning models.
Our results, evaluated on a real-world dataset collected from our institution, show that the best-performing meta-learning models outperform VLLMs in 2-way classification tasks. In contrast, in more complex 3-way classification tasks, the best-performing VLLMs slightly outperform the meta-learning models. 
While VLLMs show promising results, their reliability and practical applicability remain uncertain and require further investigation.


\end{abstract}

}

\maketitle

\IEEEdisplaynontitleabstractindextext

%
\IEEEpeerreviewmaketitle


\section{Introduction}\label{section:introduction}
\IEEEPARstart{A}{s} online education has gained popularity, the need for efficient and scalable methods of automatically grading and assessing student work has become increasingly important. Automated grading offers several advantages, including scalability, time efficiency, grading consistency, and immediate feedback.

Early research on automated grading primarily focused on closed-ended questions, such as multiple-choice and fill-in-the-blank questions, where responses could be easily verified using rule-based systems \cite{corbett1994knowledge, callear2001caa}. While these approaches provide benefits such as ease of implementation, accuracy, and speed, they are often insufficient for assessing higher-order thinking skills. Open-ended questions, on the other hand, provide deeper insight into students' reasoning and problem-solving processes, making them a more effective tool for evaluating student understanding.

\subsection{NLP in Automated Grading} 

To facilitate autograding of (semi) open-ended, particularly in textual formats such as essays and short answers, Machine Learning (ML) and, more specifically, Natural Language Processing (NLP) have been widely used \cite{sakaguchi2015effective, magooda2016vector, zhang2022automatic, erickson2020automated}. These approaches typically involve representing student responses as numerical vectors using NLP techniques and then applying a classifier or clustering to predict a score.

Various text representation techniques have been explored for this purpose. Early methods relied on handcrafted features such as Bag-of-Words (BoW) (Refs. \cite{erickson2020automated, lan2015mathematical}). Other approaches used word embeddings such as Word2Vec \cite{mikolov2013efficient} and GloVe \cite{pennington2014glove} (Refs. \cite{magooda2016vector, zhang2022automatic}). More recently, contextualised embeddings from pre-trained transformer-based models such as BERT \cite{devlin2018bert} and SBERT \cite{reimers2019sentence} have significantly improved grading accuracy (Refs. \cite{sung2019pre,condor2020exploring, baral2021improving}).

Instead of directly predicting grades, some works have taken an unsupervised approach, using similarity metrics to infer scores based on comparisons with already graded student responses (Ref. \cite{baral2021improving}).

\subsection{Challenges in Mathematical Education}

Despite advances in NLP-based autograding, assessing mathematics-related responses presents additional challenges. Unlike purely textual responses, student answers in math-related courses often contain a mix of text and mathematical expressions, including symbols, formulas, and equations. While NLP methods have been adapted to handle such responses \cite{erickson2020automated, baral2021improving}, most pre-trained NLP embedding models are not specifically trained on mathematical data, leading to poor performance. To mitigate this issue, MathBERT \cite{shen2021mathbert} was introduced, a BERT model pre-trained on a large mathematical corpus to improve autograding accuracy in this domain.

Mathematical responses in image format, which contain text and mathematical expressions, can be processed using Optical Character Recognition (OCR)~\cite{hamad2016detailed,zhang2017watch,zhang2018track}. These extracted responses can then be analyzed using NLP-based autograding frameworks. Recent studies have also explored Vision-Large Language Models (VLLMs) for assessing such responses (Refs. \cite{liu2024ai,kortemeyer2023toward}). However, research on whether VLLMs can achieve comparable performance to models specifically trained on mathematical data remains limited.

Beyond text and mathematical expressions, handwritten mathematical responses can include tables, charts, and graphs, posing additional challenges. These visual elements cannot be directly processed using traditional NLP or OCR methods and very few studies have investigated autograding for such image-based responses (Refs. \cite{baral2025drawedumath,baral2023auto}), highlighting that further research is required to develop grading models for such visual responses.

\subsection{Our Contribution}

This research focuses on the automated grading of handwritten student responses in university-level math-related courses in the format of image, that exclusively contain graphs (and text), addressing the gap in the existing literature. Our primary goal is to develop deep meta-learning models capable of automatically grading and providing feedback on students' graphs in mathematics-related education, offering teachers an objective and consistent second opinion.

Additionally, we compare the performance of specially trained meta-learning models against VLLMs to assess their effectiveness.

One major challenge in developing such a deep learning system for an assistive educational application is the lack of a labeled dataset of handwritten mathematical graphs. To address this, we utilize a dataset from a distance university in Switzerland.

We propose and address the following research questions in this study:
\begin{itemize}
    \item \textbf{RQ1}: What is an effective data preprocessing pipeline that (semi) automatically extracts the relevant information (graphs and text) from hand-drawn student responses?
    \item \textbf{RQ2}: How well do deep meta-learning models perform on this dataset?
    \item \textbf{RQ3}: How do VLLMs compare to specially trained meta-learning models for this task?
\end{itemize}

\section{Related Works}\label{section:related_works}



Compared to other educational domains, automated grading of open-ended responses in mathematics, poses more challenges. In addition to pure texts, student responses in mathematics often include mathematical expressions (e.g., symbols, formulas, and equations). 
Some previous studies such as Ref.~\cite{lan2015mathematical} focused exclusively on mathematical expressions, extracting them from student responses, parsing them with SymPy \cite{10.7717/peerj-cs.103}, and using BoW representations followed by clustering based on similarity to predict the grade. 
Erickson et al. \cite{erickson2020automated} considered both text and mathematical expressions, formulating the problem as a classification task. They experimented with BoW representations followed by Random Forest and XGBoost classifiers to predict grades. Additionally, they explored a deep learning approach using Stanford tokenization, GloVe word embeddings, and an LSTM model for grade prediction.

Baral et al. \cite{baral2021improving} built upon \cite{erickson2020automated} by introducing SBERT-Canberra model, an unsupervised model leveraging contextual embeddings from Sentence-BERT (SBERT). Instead of directly predicting the grades, their unsupervised approach used vector similarity between student responses and previously graded answers to infer scores. Later, Zhang et al. \cite{zhang2022automaticmeta} improved upon \cite{baral2021improving} by incorporating MathBERT, a BERT model pre-trained on mathematical text. Additionally, they employed in-context learning, where MathBERT was provided with the question, student response, and a few labeled examples before fine-tuning the model for automated grading.

While these studies focused on text-based responses, student answers can also be submitted as handwritten images containing additional elements such as diagrams, tables, charts, and graphs, adding further complexity to the autograding process. Baral et al. \cite{baral2023auto} extended their previous work \cite{baral2021improving} by incorporating CLIP \cite{radford2021learning} to encode images alongside SBERT representations. Although their approach did not significantly improve state-of-the-art results,
it demonstrated a new research direction toward integrating vision-based models in autograding.

Recently, there has been growing interest in leveraging (V)LLMs for autograding. Their extensive training on diverse datasets—including textual content, mathematical expressions, codes, and images—along with their reasoning capabilities, makes them promising tools for automated assessment. Caraeni et al. \cite{caraeni2024evaluating} evaluated GPT-4o’s ability to grade scanned handwritten responses in a university-level probability theory exam. They compared different prompting strategies, finding that GPT-4o performed best when provided with the student’s response, the correct answer, and grading rubrics. Similarly, Liu et al. \cite{liu2024ai} assessed GPT-4’s performance in grading university-level mathematics exams, using OCR to extract text and equations before prompting the model with student responses, grading rubrics, and maximum scores. In the context of physics education, Kortemeyer et al. \cite{kortemeyer2023toward} leveraged Mathpix OCR to convert handwritten student responses into LaTeX before applying LLMs for automated grading.

Baral et al. \cite{baral2025drawedumath} refined a portion of the K-12 math dataset, collected initially from the online learning platform ASSISTments \cite{heffernan2014assistments}, by hiring teachers to annotate student answers with detailed descriptions and question-answer pairs. They then evaluated multiple VLLMs on answering questions about student responses, concluding that current models still have significant room for improvement in educational applications.

While these studies have explored automated grading of handwritten responses mainly containing text, mathematical expressions, tables, and charts,
our work focuses specifically on the automated assessment of hand-drawn graphs in mathematical education, a largely unexplored area. We investigate deep learning models for grading such responses, aiming to provide objective and consistent feedback for educators.

\section{Experiments}\label{section:methods}

\subsection{Data Collection}

The dataset used in this study consists of anonymized images of hand-drawn graphs collected in compliance with relevant data protection regulations, from the Moodle online learning platform at a Swiss distance university of applied sciences.

Students created these graphs using a digital canvas integrated into the Moodle platform as part of online mathematics-related courses, specifically in economics (Volkswirtschaftslehre in German). These images contain graphs drawn in response to quiz/exam questions and assignments requiring graphical solutions. Some students also included textual explanations within the images. To facilitate the drawing process, the institution provided students with Wacom tablets beforehand.

The dataset initially consists of approximately 1,372 images collected from three different course modules: VWL7 (HS21/22), VWL8 (FS22), and VWL9 (FS22). Each module comprises multiple tasks or assignments, with the number of participating students ranging from 60 to 90. After filtering out empty responses and irrelevant drawings, 1,174 images remained in the final dataset.

Several challenges will be addressed in this study, including missing labels (grades) and feedback, variations in scale and axis alignment of the graphs, the presence of text or unnecessary margins within the images, and the inherent difficulties posed by a small and imbalanced dataset. Addressing these issues is crucial for developing a robust autograding system.

\subsection{Data Annotation}

Creating appropriate labels for the graphs is the initial step toward training the model. The only available labels were incomplete grades for the respective graphs, without any accompanying grading criteria or feedback. Since these grades were assigned by different teachers without standardized rubrics, they lack consistency and are not particularly useful.

To address this, we collaborated with an expert teacher to annotate the graphs systematically. The annotation process involves two key steps: 1) Defining evaluation criteria (rubrics) for each task in each module and 2) Applying these grading criteria to annotate the graphs, indicating which criteria are met in each graph.
To facilitate this process, we developed a dedicated web application for annotation. The expert annotator uses this platform to create grading criteria and systematically assess the graphs based on the predefined grading criteria.

An example of a task description and its corresponding grading criteria, created by the expert annotator, is shown in Table~\ref{tab:task_description}. In this example, the assignment includes a single grading criterion, which can either be fulfilled or not in the submitted graphs.

\begin{table}[ht]
    \caption{A sample of task description and grading criteria created by the expert annotator.}
    \label{tab:task_description}
    \scriptsize
    \centering
    \begin{tabular}{|p{8cm}|}
        \hline 
        \\
        \texttt{Unten folgend wird eine Situation beschrieben. Welche Auswirkungen hat diese Situationen auf einem perfekten Markt (=vollkommene Konkurrenz)?} \\
        \\
        \texttt{Zeichnen Sie ein Preis-Mengen-Diagramm, beschriften Sie dessen Achsen und die beiden Kurven und bezeichnen Sie das Gleichgewicht. Zeichnen Sie dann die neue Situation ein und zeigen die Auswirkungen auf Preis und Menge. Wählen Sie jene Auswirkung, welche unmittelbar auf die Situation folgt. Beachten Sie, dass alle übrigen Bedingungen gleich bleiben.}\\
        \\
        \textbf{Situation: }\texttt{Der herannahende Wirbelsturm "Sandy" an der Ostküste der USA hatte viele Bewohner dazu bewegt, sich mit lebensnotwendigen Gütern einzudecken und Häuser \& Wohnungen behelfsmässig gegen Zerstörung zu sichern.} \\
        \\
        \textbf{Betrachteter Markt: }\texttt{Markt für Bretter (Einsatz zur Fenster- und Türsicherung)} \\
        \\
        \hline\hline
        \\
        \textbf{Grading Criteria: }\texttt{[correct shift of demand curve to the right]} \\
        \\
        \hline
    \end{tabular}
    
\end{table}

The translation of the task description is: 
\textit{
Below, a situation is described. What effects does this situation have on a perfect market (i.e., perfect competition)?
Draw a price-quantity diagram, label its axes and the two curves, and mark the equilibrium. Then, depict the new situation and show the impact on price and quantity. Choose the effect that follows immediately from the situation. Assume that all other conditions remain unchanged. \\
Situation: The approaching hurricane "Sandy" on the East Coast of the USA prompted many residents to stock up on essential goods and secure their houses and apartments temporarily against destruction.\\
Market Considered: Market for planks (used for window and door reinforcement)
}

\subsection{Data Preprocessing}
The collected images contain not only graphs, but also relevant text and unnecessary margins. To extract the graphs, we apply traditional computer vision techniques, including morphological transformations and contour detection. The graph is identified as the largest square-like contour. However, since other similar contours may be mistakenly detected as graphs, the final step prompts the annotator to confirm the correct selection of the bounding box for the graph. If necessary, the annotator can manually adjust the bounding rectangle, though such interventions are very rare.

Since text within the graphs may be relevant to the solution, we also use Tesseract OCR \cite{smith2007overview} to extract text from images. Both the extracted graph and text will serve as inputs for the machine learning model.

For grading, we define a grading system based on the fulfillment of criteria annotated by the expert annotator. The grade for each graph is computed as:

\begin{equation}\label{eq:calculate-grade}
\text{Grade} = \sum_{i=0}^{n-1} C_i \cdot 2^i 
\end{equation}

where $n$ represents the total number of grading criteria for a given assignment, and $C_i$ is a binary indicator, 1 if criterion $i$ is met, 0 otherwise. This encoding ensures that each unique combination of fulfilled criteria corresponds to a distinct numerical grade. For example, if an assignment has a single grading criterion, the possible grades are 0 and 1. If there are two grading criteria, the possible grades expand to 0, 1, 2, and 3.

These numerical grades serve as the target labels for model training.

\subsection{Problem Formulation and Evaluation Metrics}

Given a dataset of student-drawn images, each consisting of a graph and extracted text, our objective is to develop a machine learning model capable of grading these graphs and providing criterion-specific feedback. The primary challenge is learning from few labeled examples per module assignment, making this a few-shot learning problem.

Formally, let \( \mathcal{D} = \{(X_i, y_i)\}_{i=1}^{N} \) denote the dataset, where each sample consists of:
\begin{itemize}
    \item \( X_i = (G_i, T_i) \), where \( G_i \) represents the hand-drawn graph and \( T_i \) is the extracted text from the image.
    \item \( y_i \) is the grade assigned to the graph based on \(\{C_1^i, C_2^i, ..., C_m^i\} \), binary vector indicating whether each of the \( m \) predefined grading criteria is fulfilled (\( C_j^i = 1 \) if fulfilled, 0 otherwise).

\end{itemize}

The model performance is evaluated based on the grading accuracy, agreement between predicted grades \( \hat{y}_i \) and ground-truth grades \( y_i \). This ensures that all (un)fulfilled criteria in a given prediction match those in the ground-truth.




\subsection{Meta-learning}

Deep learning has achieved remarkable success in image classification tasks \cite{krizhevsky2012imagenet,zeiler2014visualizing,simonyan2015very,szegedy2015going,he2016deep}. However, deep learning models typically require large amounts of labeled data and extensive training to achieve high performance. In many real-world applications, including ours, labeled data is scarce, and only a few examples are available for each class. This limitation makes it challenging to train traditional deep learning models effectively.

Meta-learning, also known as learning to learn or few-shot learning, addresses this challenge by enabling models to generalize to new tasks and unseen classes with only a few labeled examples. Instead of training a specific task with a large number of samples per class, meta-learning frameworks are designed to extract transferable knowledge from multiple tasks, allowing the model to quickly adapt to new tasks with minimal data. Meta-learning is performed in an episodic manner, where each episode simulates a few-shot learning scenario. In each episode, $N$ classes are randomly selected, and for each class, $K$ examples are sampled (known as $N$-way $K$-shot) to form the support set. A separate query set is constructed using additional examples from the same $N$ classes. The support set is used to learn, while the query set evaluates how well the model generalizes within the episode.
Depending on the meta-learning approach, the algorithm utilizes the support set in different ways: 1) Metric-based methods define a class representation (e.g., class prototypes in prototypical networks) or learn a similarity function (e.g., relation networks) that compares support and query samples to classify the queries; 2) Optimization-based methods use the support set to perform a few gradient-descent updates on the model’s parameters before evaluating on the query set; and 3) Model-based methods leverage the support set to update an internal memory (e.g., RNN's activations), which is then used to evaluate the query set.
During meta-learning, the model is trained to minimize the error between the predicted labels and the ground-truth labels for the query set.

In our work, in each episode we sample classes from the same module assignment. This way the focus of the model is to learn the task itself rather than to learn to which modules and assignments the graphs correspond. 
In this study, we evaluate the performance of the following meta-learning algorithms on our dataset: Matching Network, Prototypical Network, Relation Network, MAML, and ProtoMAML. We assume an \( N \)-way \( K \)-shot setting, where the support set is denoted as \( S = \{(X_1, y_1), (X_2, y_2), \dots, (X_{KN}, y_{KN})\} \), with \( S_k \) representing the subset of \( S \) containing examples with label \( k \). The query set is denoted as \( Q \), and \( g_\theta \) is a function that embeds both the graph and the extracted text, combining them through concatenation of their embedding vectors. The parameters of which are learned during meta learning. In the following, we provide a brief summary of each model used in this study. For a complete description, the reader is referred to the original publications.

\subsubsection{Matching Network}

Proposed by Vinyals et al. \cite{vinyals2016matching}, Matching Network uses an attention mechanism to classify query examples based on their similarity to support set examples. The probability of a query example \( X \) belonging to class \( k \) is given by:

\begin{equation}
    p(\hat{y} = k | X) = \sum_{(X_i,y_i)\in S} \alpha(X, X_i) \mathds{1}(y_i = k)
\end{equation}

where \( \mathds{1}(y_i = k) \) is an indicator function that equals 1 if \( y_i = k \) and 0 otherwise, and \( \alpha \) is the normalized attention weight defined as:

\begin{equation}
\alpha(X, X_i) = \frac{\exp(g_\theta(X)^T g_\theta(X_i))}{\sum_j \exp(g_\theta(X)^T g_\theta(X_j))}
\end{equation}

During meta-learning, the model minimizes the negative log-likelihood loss, \( -\log p(\hat{y} | X) \), over the query set to find the optimal embedding function parameters.

\subsubsection{Prototypical Network}  
Introduced by Snell et al. \cite{snell2017prototypical}, the Prototypical Network represents each class by a prototype, computed as the mean of the embedded support set examples:  
\begin{equation}
    c_k =\frac{1}{|S_k|} \sum_{(X_i,y_i)\in S_k} g_\theta(X_i).
\end{equation}  

Query examples are classified based on their distance to these prototypes. The probability of a query example \( X \) belonging to class \( k \) is given by:  
\begin{equation}
    p(\hat{y}=k|X) = \frac{\exp(-d(g_\theta(X),c_k))}{\sum_{k'}\exp(-d(g_\theta(X),c_{k'}))},
\end{equation}  
where \( d(g_\theta(X), c_k) \) is the Euclidean distance between the query embedding \( g_\theta(X) \) and the prototype \( c_k \). During meta-learning, the model minimizes a similar negative log-likelihood loss as in the Matching Network, \( -\log p(\hat{y}=k|X) \), to optimize the embedding function parameters.

\subsubsection{Relation Network}  

Unlike simple similarity-based methods such as Matching Networks and Prototypical Networks, the Relation Network, proposed by Sung et al. \cite{sung2018learning}, learns a metric function \( f_\phi \) to determine the similarity between support and query examples.
Instead of relying on predefined distance metrics,  
it models similarity through a learnable relation module. 
The relation module takes as input the embeddings of a support-query pair, typically concatenated, and outputs a relation score between 0 and 1, with 1 indicating a match and 0 a mismatch:
\begin{equation}
    r_{ij} = f_\phi\left( \text{Concat}\left(g_\theta(X_i), g_\theta(X_j)\right) \right),
\end{equation}  
where \( X_i \) and \( X_j \) correspond to support and query set examples, respectively.  
During meta-learning, the model is trained by minimizing a mean squared error (MSE) loss between the predicted relation scores and the ground truth labels.  

\subsubsection{MAML}  

Model-Agnostic Meta-Learning (MAML), introduced by Finn et al. \cite{finn2017model}, aims to learn a model initialization that can be quickly adapted to new tasks with a few gradient updates. In each episode \(T_i\), MAML performs a few steps of gradient descent on the support set to compute the updated parameters, \(\theta'\):  
\begin{equation}
    \theta'_i = \theta - \alpha \nabla_\theta L_{T_i}(\theta, S),
\end{equation}  
where \( L_{T_i} \) is the loss function for task \( T_i \). Here we assume that the model contains the mebdding function as well as a classifier (e.g. fully connected layer on top of it) and \(\theta\) refers to all parameters including the ones from the classifier. The updated parameters are then used to evaluate the model on the query set. During meta learning the parameters are updated by minimizing the query loss across all episodes:  
\begin{equation}
    \theta \leftarrow \theta - \beta \nabla_\theta \sum_{T_i} L_{T_i}(\theta'_i, Q),
\end{equation}  
To avoid the computational expense of second-order derivatives, we use First-Order MAML (FOMAML), which approximates the gradients by ignoring the second-order terms.

\subsubsection{ProtoMAML}
Proposed by Triantafillou et. al. \cite{Triantafillou2020Meta-Dataset:}, ProtoMAML is a hybrid meta-learning algorithm that combines the fast adaptation of MAML with the inductive bias of the Prototypical Network to improve convergence in few-shot learning. Instead of initializing classifier parameters randomly, ProtoMAML uses the class prototypes as the initial weights, setting them as follows:
\begin{align}
    W_k &= 2c_k \notag \\
    b_k &= -||c_k||^2
\end{align}
After this initialization, the rest of the procedure follows the same approach as MAML.\\

\subsection{VLLMs}
(V)LLMs have demonstrated competitive performance in various tasks, including automatic
grading, often rivaling models specifically trained for such
tasks. To assess the performance of VLLMs and compare it with our specifically trained meta-learning models, we also evaluate few-shot in-context learning with VLLMs. Unlike meta-learning, which trains a model to adapt quickly with minimal examples, in-context learning does not require fine-tuning. Instead, the model is provided with a few labeled examples (support set) as part of its input prompt and is expected to classify new queries based on learned patterns. 


\section{Results}\label{section:results}

In this section, we present the results that correspond to our research questions. To address \textbf{RQ1}, we first describe a semi-automatic data preprocessing pipeline to extract hand-drawn graphs and text from student responses (Section~\ref{section:result-data-preprocessing}). We then address \textbf{RQ2} by evaluating the performance of deep meta-learning models on our dataset (Section~\ref{section:result-meta-learning}) and further analyze the contribution of each input modality in Section~\ref{section:result-ablation-study}. Finally, to address \textbf{RQ3}, we compare the grading performance of VLLMs with that of the specialized meta-learning models (Section~\ref{section:result-vllms}).


\subsection{Data Preprocessing}\label{section:result-data-preprocessing} 
\begin{figure*}[!h]
    \centering
    \includegraphics[width=\textwidth]{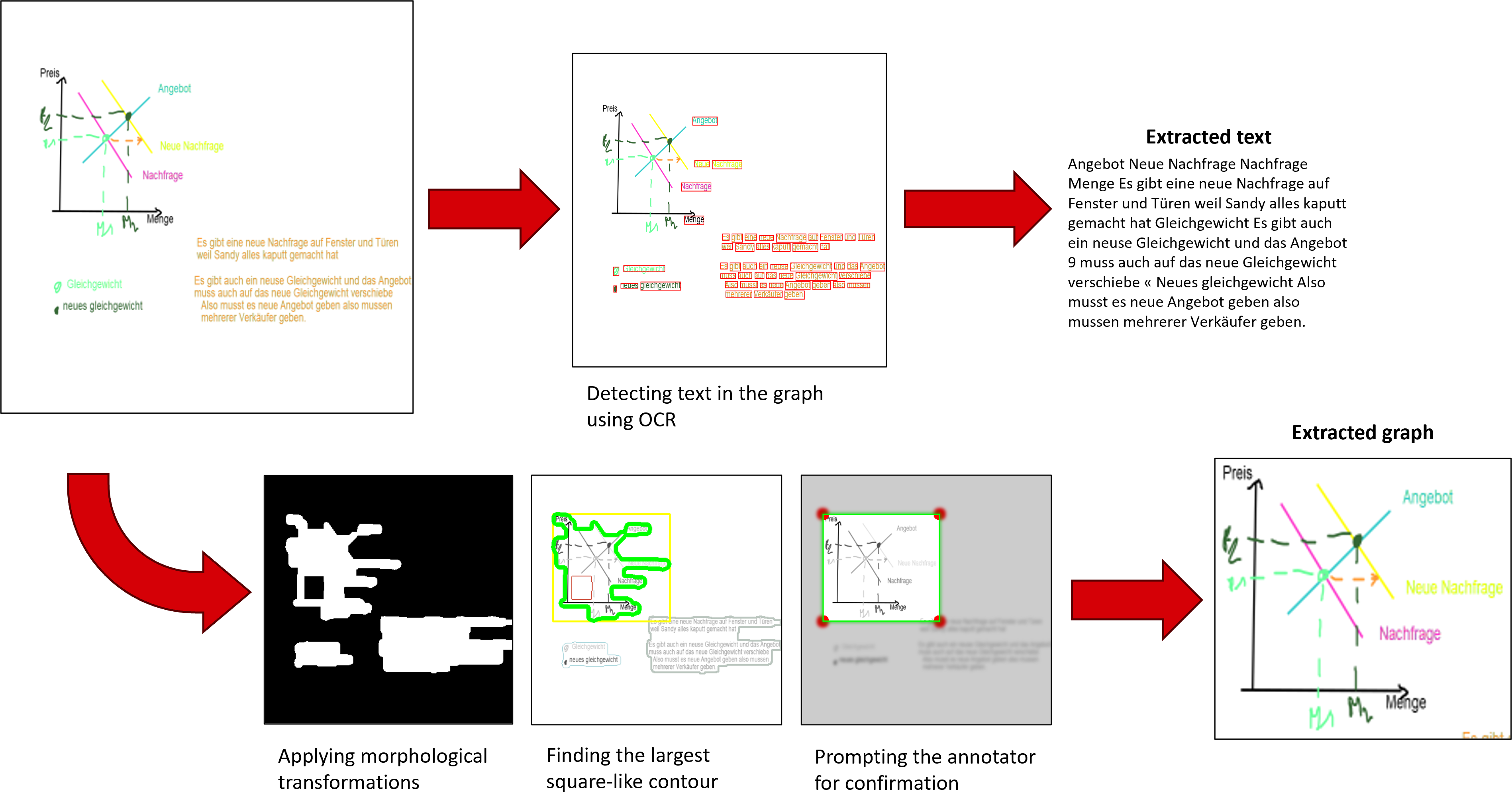}
    \caption{Data preprocessing pipeline for an image. (Left) The original student-drawn image containing graph, text, and extraneous margins. (Top) OCR is applied to extract text. (Bottom) Morphological transformations followed by contour detection are used to identify the graph (largest square-like shape, highlighted in green, with its enclosing rectangle in yellow). The extracted graph is finalized with a verification step allowing manual adjustment if necessary.}
    \label{fig:data_preprocessing}
\end{figure*}

Our data preprocessing pipeline was designed to isolate the relevant graph regions from each student-submitted image and extract any accompanying textual information.
Figure~\ref{fig:data_preprocessing} shows the process for a sample image (corresponding to the given task in Table~\ref{tab:task_description}). In this figure, the detected contour and its enclosing bounding rectangle are highlighted to demonstrate the region of interest identified during processing.

The extracted graphs were all resized to \((224,224)\). To preserve meaningful visual information for model training, we retain the original colors in the graphs. Additionally, during training, we apply data augmentation techniques such as adding small random rotations and perspective transformations to improve the robustness of the model and partially address the limited dataset size.

Table~\ref{tab:stats} provides a detailed breakdown of the dataset, showing the number of graphs for each module, assignment, and grade combination. The dataset consists of three modules: VWL7 (HS21/22), VWL8 (FS22), and VWL9 (FS22). The number of assignments per module varies: VWL7 contains 8 assignments, VWL8 has 4, and VWL9 includes 5. The number of grading criteria per assignment also varies, with some assignments allowing only two possible grades (e.g., 0 and 1), while others have four possible grades (0, 1, 2, and 3), indicating the presence of two grading criteria. It is important to note that while some assignments, such as VWL9 assignment 2, exhibit a balanced distribution with all four grades (0, 1, 2, 3) well represented, others mostly show a highly imbalanced distribution. For instance, in VWL7 assignment 6, grade 3 appears in 52 instances, whereas grades 1 and 2 are significantly underrepresented, with only 3 and 1 instances, respectively. This imbalance further complicates the construction of meta-learning episodes and limits model generalization.

To construct valid few-shot classification tasks, we must ensure that enough labeled examples exist per grade category within each assignment. Specifically, an N-way K-shot meta-learning task requires N distinct grades (classes), each with K support examples and at least one query example. Thus, the minimum number of samples required for an N-way K-shot task is $N\times(K+1)$. For example, a 2-way 1-shot task requires at least 4 examples—2 distinct grades with 1 support and 1 query example each—from a single assignment. Similarly, a 3-way 4-shot task would require a minimum of 15 examples distributed across three different grades within the same assignment. Since our dataset is limited in size, the number of assignments that meet these criteria decreases rapidly as N and K increase. To visualize this limitation, we highlight in Table~\ref{tab:stats} the rows corresponding to grade categories with only one instance. These rows are insufficient even for the simplest 1-shot setting, making them unusable in meta-learning tasks. This illustrates the sparsity of usable data at higher K-shot levels. Consequently, we restricted our experiments to 2-way and 3-way classification with up to 4-shot learning.


\begin{table*}[!h]
\centering
\caption{The number of graphs in each (module, assignment, grade) combination.}
\label{tab:stats}
\begin{tabular}{|>{\centering\arraybackslash}p{3cm}|>{\centering\arraybackslash}p{2cm}|>{\centering\arraybackslash}p{1.5cm}|>{\centering\arraybackslash}p{1.5cm}|}
\toprule
Module & Assignment & Grade & Number of graphs \\
\midrule
\multirow{26}{*}{VWL7-HS21/22} & \multirow{4}{*}{1} & 0 & 3 \\
 & & \cellcolor{lightgray}1 & \cellcolor{lightgray}1 \\
 & & 2 & 10 \\
 & & 3 & 30 \\\cline{2-4}
 & \multirow{2}{*}{2} & 0 & 45 \\
 & & 1 & 20 \\\cline{2-4}
 & \multirow{2}{*}{3} & 0 & 24 \\
 & & 1 & 40 \\\cline{2-4}
 & \multirow{2}{*}{4} & 0 & 9 \\
 & & 1 & 54 \\\cline{2-4}
 & \multirow{4}{*}{5} & 0 & 7 \\
 & & 1 & 35 \\
 & & \cellcolor{lightgray}2 & \cellcolor{lightgray}1 \\
 & & 3 & 18 \\\cline{2-4}
 & \multirow{4}{*}{6} & 0 & 5 \\
 & & 1 & 3 \\
 & & \cellcolor{lightgray}2 & \cellcolor{lightgray}1 \\
 & & 3 & 52 \\\cline{2-4}
 &  \multirow{4}{*}{7} & 0 & 20 \\
 & & \cellcolor{lightgray}1 & \cellcolor{lightgray}0 \\
 & & 2 & 21 \\
 & & 3 & 17 \\\cline{2-4}
 &  \multirow{4}{*}{8} & 0 & 4 \\
 & & \cellcolor{lightgray}1 & \cellcolor{lightgray}1 \\
 & & \cellcolor{lightgray}2 & \cellcolor{lightgray}1 \\
 & & 3 & 53 \\
\hline
\multirow{12}{*}{VWL8-FS22} & \multirow{2}{*}{1} & 0 & 7 \\
 & & 1 & 58 \\\cline{2-4}
 & \multirow{4}{*}{17} & 0 & 7 \\
 & & 1 & 23 \\
 & & \cellcolor{lightgray}2 & \cellcolor{lightgray}0 \\
 & & 3 & 32 \\\cline{2-4}
 & \multirow{4}{*}{3} & 0 & 17 \\
 & & 1 & 5 \\
 & & 2 & 37 \\
 & & 3 & 7 \\\cline{2-4}
 & \multirow{2}{*}{4} & 0 & 20 \\
 & & 1 & 36 \\
\hline
\multirow{18}{*}{VWL9-FS22} & \multirow{4}{*}{1} & 0 & 17 \\
 & & 1 & 52 \\
 & & 2 & 4 \\
 & & 3 & 19 \\\cline{2-4}
 & \multirow{4}{*}{2} & 0 & 32 \\
 & & 1 & 50 \\
 & & 2 & 4 \\
 & & 3 & 7 \\\cline{2-4}
 & \multirow{4}{*}{3} & 0 & 50 \\
& & \cellcolor{lightgray}1 & \cellcolor{lightgray}1 \\
 & & 2 & 2 \\
 & & 3 & 32 \\\cline{2-4}
 & \multirow{4}{*}{5} & 0 & 11 \\
 & & 1 & 11 \\
 & & \cellcolor{lightgray}2 & \cellcolor{lightgray}1 \\
 & & 3 & 69 \\\cline{2-4}
 & \multirow{2}{*}{8} & 0 & 64 \\
 & & 3 & 24 \\
\bottomrule
\end{tabular}
\end{table*}


\subsection{Meta-Learning}\label{section:result-meta-learning} We evaluated several meta-learning algorithms for each N-way, K-shot tasks. For graph embeddings, we employed a ResNet-18 architecture, and for text embeddings, we used a BERT model pre-trained on German (uncased). Both networks were fine-tuned during the meta-learning, ensuring that the representations are adapted to the few-shot learning task. All models were trained for 1,000 epochs on a GPU, with 100 inner gradient steps applied for the FOMAML and ProtoFOMAML methods.

Table~\ref{tab:updated-meta_learning_results} presents the mean accuracy (with standard deviation) across 200 episodes of five meta-learning approaches---Prototypical Network, FOMAML, ProtoFOMAML, Matching Network, and Relation Network---evaluated in both 2-way and 3-way settings under 1-, 2-, and 4-shot configurations. 
In the 2-way tasks, the ProtoFOMAML achieved the best accuracy (54.09\% $\pm$ 2.83) in 1-shot while Prototypical Network yielded the highest accuracy in 2-shot (55.76\% $\pm$ 0.01) and 4-shots (56.87\% $\pm$ 3.62). 
For the more challenging 3-way tasks, Matching Network performed best in 1-shot (42.73$\% \pm$ 4.56) while FOMAML outperformed other models in 2-shot (41.65\% $\pm$ 3.25) and 4-shot (41.30\% $\pm$ 5.21) settings. Overall, the observed drop in accuracy when moving from 2-way to 3-way tasks confirms the increased difficulty of distinguishing among more classes.


\begin{table*}[ht]
    \centering
    \caption{Mean accuracy (\%) with standard deviation for different meta-learning models, N-Way, and K-Shot settings.}
    \label{tab:updated-meta_learning_results}
    \begin{tabular}{ccccccc}
        \toprule
        N-way & K-shot & PrototypicalNetwork & FOMAML & ProtoFOMAML & MatchingNetwork & RelationNetwork \\
        \midrule
        \multirow{3}{*}{2} & 1 & 51.43 ± 3.07 & 50.73 ± 2.66 & \textbf{54.09 ± 2.83} & 50.32 ± 2.98 & 52.07 ± 2.89 \\
        & 2 & \textbf{55.76 ± 3.02} & 51.57 ± 2.72 & 53.49 ± 3.01 & 54.29 ± 3.16 & 53.13 ± 3.12 \\
        & 4 & \textbf{56.89 ± 3.61} & 52.46 ± 3.44 & 55.93 ± 3.73 & 52.53 ± 3.67 & 55.44 ± 3.64 \\
        \midrule
        \multirow{3}{*}{3} & 1 & 35.57 ± 4.34 & 35.15 ± 3.51 & 34.68 ± 3.71 & \textbf{42.69 ± 4.55} & 34.70 ± 4.03 \\
        & 2 & 36.06 ± 3.83 & \textbf{41.65 ± 3.25} & 36.66 ± 3.81 & 37.04 ± 3.80 & 38.57 ± 3.73 \\
        & 4 & 41.19 ± 4.89 & \textbf{41.26 ± 5.21} & 33.42 ± 4.27 & 39.07 ± 4.64 & 32.43 ± 4.60 \\
        \bottomrule
    \end{tabular}
\end{table*}

\begin{figure*}
    \centering
    \includegraphics[width=0.7\linewidth]{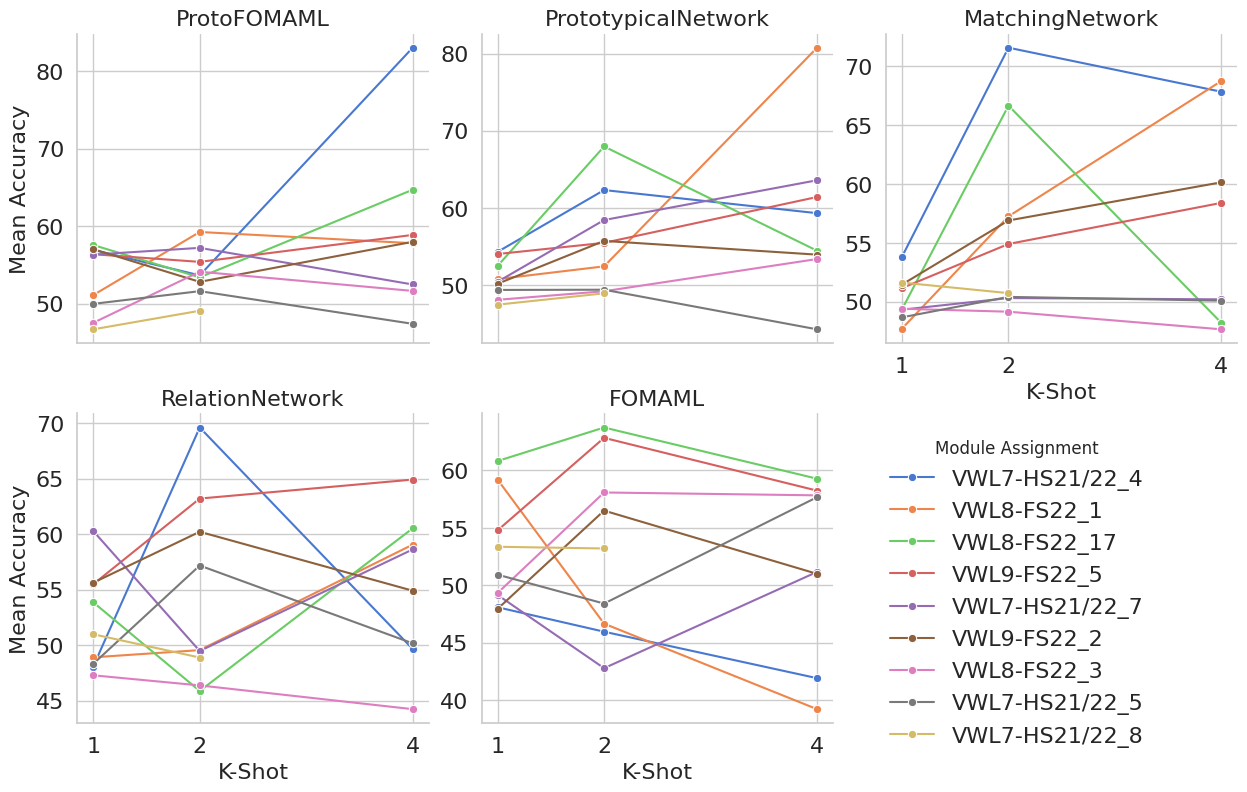}
    \caption{Performance of each meta-learning model across different module assignments for 2-way.}
    \label{fig:results_meta_by_mas_nway2}
\end{figure*}

\begin{figure*}
    \centering
    \includegraphics[width=0.7\linewidth]{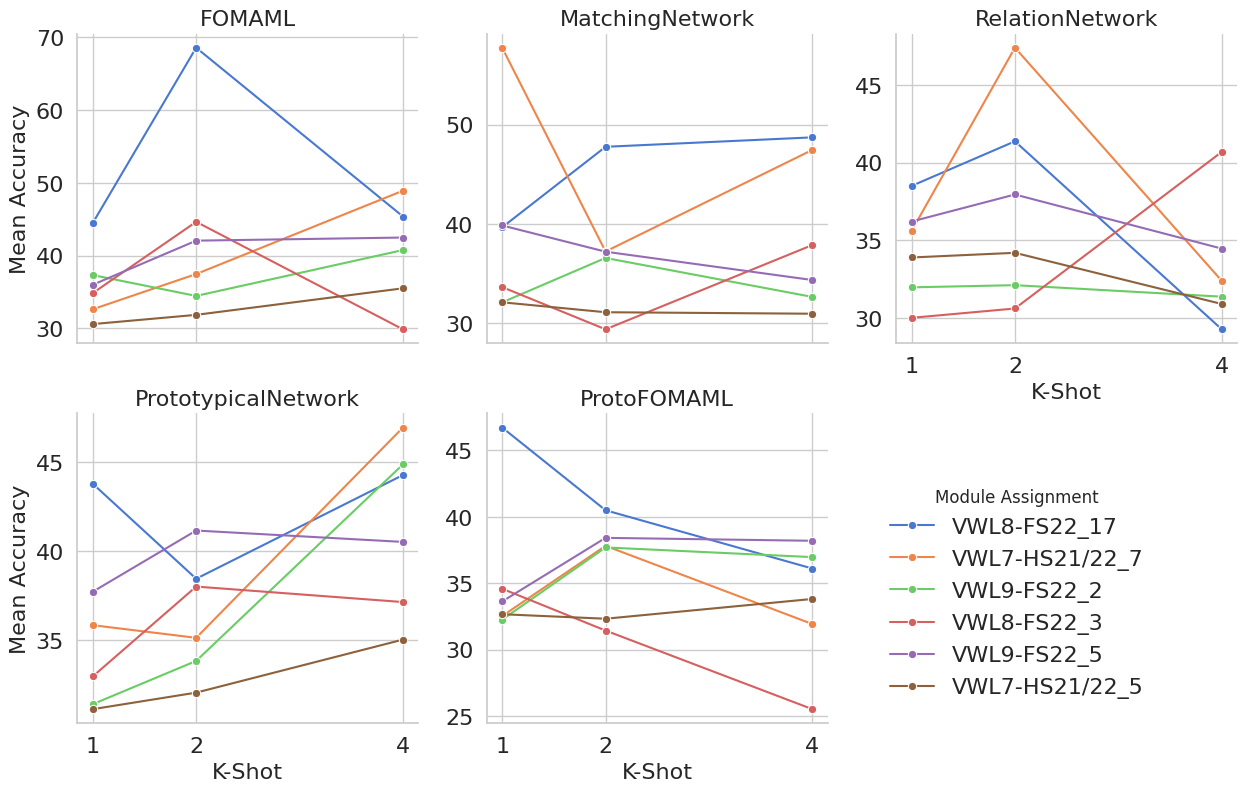}
    \caption{Performance of each meta-learning model across different module assignments for 3-way.}
    \label{fig:results_meta_by_mas_nway3}
\end{figure*}

To better understand how task content influences model performance, we evaluated models separately for each task. Each task is identified by a combination of module and assignment, formatted as Module\_Assignment (e.g., VWL7-HS21/22\_4 refers to assignment 4 in module VWL7-HS21/22).
Figures~\ref{fig:results_meta_by_mas_nway2} and~\ref{fig:results_meta_by_mas_nway3} show accuracy across different K-shot values for each model, separated by task, in the 2-way and 3-way settings respectively.
In the 2-way setting (Figure~\ref{fig:results_meta_by_mas_nway2}), some tasks exhibit consistently stronger performance (e.g., VWL7-HS21/22\_4 and VWL8-FS22\_17), while others, such as VWL7-HS21/22\_8 and VWL8-FS22\_3 show lower accuracy across most K-shot settings. This indicates that certain assignments are inherently easier or more informative for learning than others.
Similarly, in the 3-way setting (Figure~\ref{fig:results_meta_by_mas_nway3}), tasks such as VWL8-FS22\_17 consistently result in higher accuracy, while others (e.g., VWL7-HS21/22\_5) lead to lower performance. These patterns persist across different models and K-shot settings, highlighting that task difficulty varies significantly between module assignments.

\subsection{Ablation Study}\label{section:result-ablation-study}
To evaluate the individual contributions of each input modality, we conducted an ablation study on the best-performing meta-learning models in each N-way K-shot setting. The experiments compared the performance of models using only graph embeddings versus only text embeddings. These were then compared against the full multi-modal models using both input types.

The results, shown in Table~\ref{tab:updated_ablation_study_updated}, reveal two clear patterns: 1) Graph-only embeddings consistently outperformed text-only embeddings across all scenarios. This is expected since graph representations contain richer semantic and structural information about the answers, whereas text inputs largely provide complementary context. 2) Combining both modalities generally leads to higher accuracy compared to using either modality alone. In all settings except for the 2-way, 4-shot Prototypical Network, the multi-modal approach yielded a better performance.

These findings confirm that while graph representations are inherently more informative for this task, incorporating text generally strengthens the model. 



\begin{table*}[!h]
\centering
\caption{Ablation Study: Mean accuracy (\%) with standard deviation for the best-performing models with only graph embedding vs. only text embedding.}
\label{tab:updated_ablation_study_updated}
\begin{tabular}{lcc|c}
\toprule
\textbf{Model} & \textbf{\(N\)-way, \(K\)-shot} & \textbf{Only Graph Embedding} & \textbf{Only Text Embedding} \\ 
\midrule
ProtoFOMAML & 2-way, 1-shot & 52.86 ± 2.80 & 49.81 ± 2.62 \\
Prototypical Network & 2-way, 2-shot & 54.26 ± 3.14 & 48.47 ± 0.93 \\
Prototypical Network & 2-way, 4-shot & 59.33 ± 3.54 & 47.32 ± 1.18 \\
Matching Network & 3-way, 1-shot & 35.25 ± 4.09 & 33.88 ± 3.58 \\
FOMAML & 3-way, 2-shot & 40.85 ± 3.92 & 33.64 ± 3.49 \\
FOMAML & 3-way, 4-shot & 36.43 ± 4.70 & 32.63 ± 4.26 \\
\bottomrule
\end{tabular}
\end{table*}

\begin{table*}[ht]
    \centering
    \caption{Accuracy (\%) of different LLMs across various N-way, K-shot settings.}
    \label{tab:updated-llm_results}
    \begin{tabular}{cccccccc}
        \toprule
        N-way & K-shot & \texttt{gpt-4.1} & \texttt{gpt-4.1-mini} & \texttt{gpt-4.1-nano} & \texttt{gpt-4o} & \texttt{gpt-4o-mini} & \texttt{o4-mini} \\
        \midrule
        \multirow{3}{*}{2} & 1 & \textbf{43.93} & 38.19 & 29.29 & 38.46 & 29.07 & 42.39 \\
        & 2 & \textbf{44.14} & 39.62 & 43.27 & 43.80 & 31.82 & 40.50 \\
        & 4 & \textbf{53.09} & 45.16 & 50.00 & 42.05 & 46.35 & 43.23 \\
        \midrule
        \multirow{3}{*}{3} & 1 & 42.36 & 41.67 & 35.85 & \textbf{43.75} & 32.08 & 35.42 \\
        & 2 & 40.30 & 42.28 & 35.29 & \textbf{45.64} & 42.02 & 37.31 \\
        & 4 & 43.40 & 43.40 & 43.48 & 45.65 & 40.22 & \textbf{50.00} \\
        \bottomrule
    \end{tabular}
\end{table*}

\begin{figure*}[ht]
    \centering
    \includegraphics[width=0.7\linewidth]{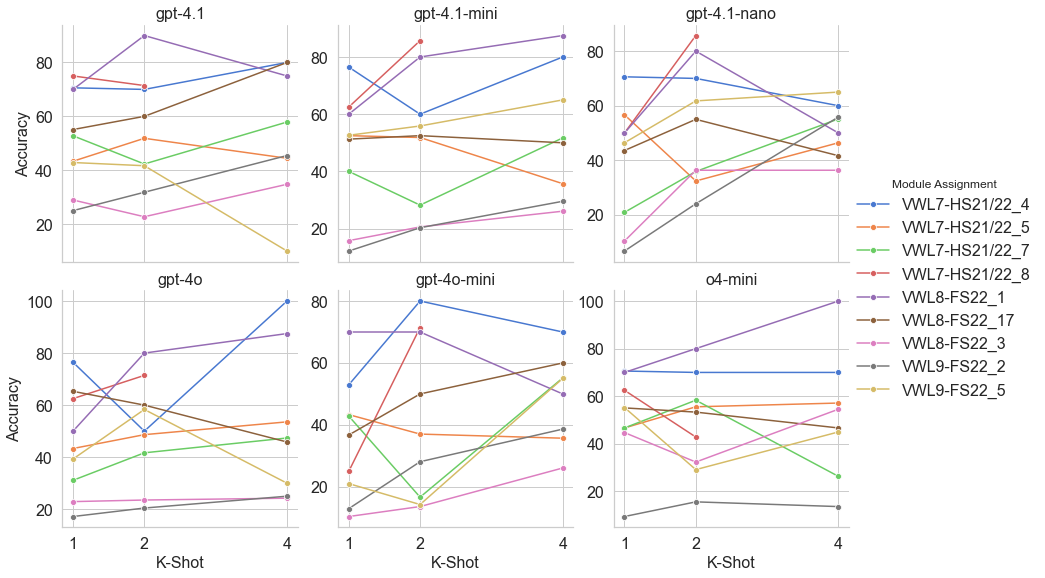}
    \caption{Performance of each LLM across different module assignments for 2-way.}
    \label{fig:results_llm_by_mas_nway2}
\end{figure*}

\begin{figure*}[ht]
    \centering
    \includegraphics[width=0.7\linewidth]{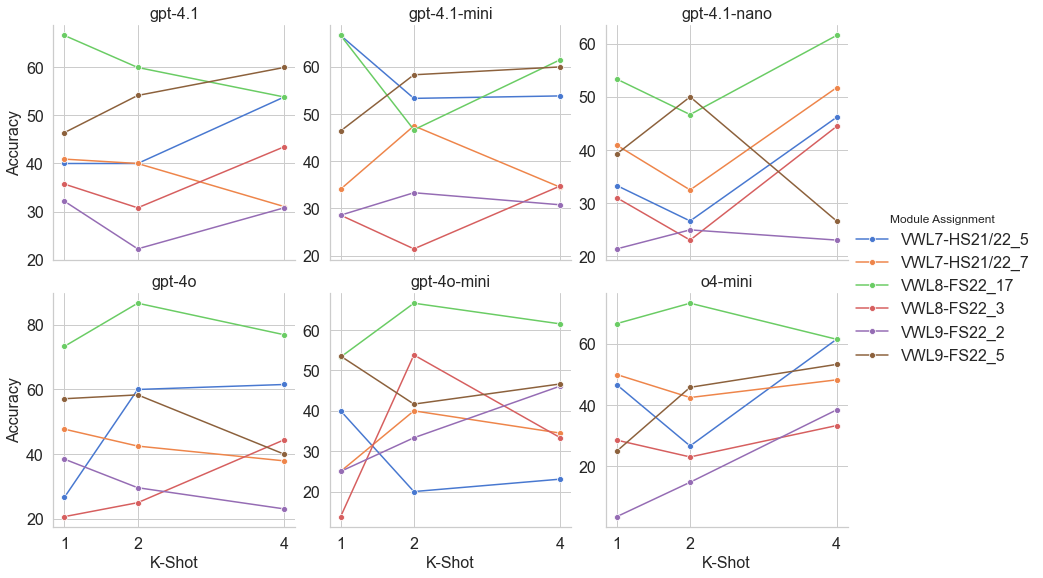}
    \caption{Performance of each LLM across different module assignments for 3-way.}
    \label{fig:results_llm_by_mas_nway3}
\end{figure*}

\begin{table}[ht]
    \caption{Example prompt for 3-way 1-shot classification with VLLMs. It includes system instructions, task description, grading criteria, and sample user-assistant interactions with Base64-encoded graphs and their expected outputs in JSON format.}
    \label{tab:llm-prompt}
    \scriptsize
    \centering
    \begin{tabular}{|p{8cm}|}
        \hline
        \\
        \textbf{System: }\texttt{You are a helpful assistant for grading students' handwritten responses in math-related economics courses. 
        You will be provided with a task description, an ordered list of grading criteria, and an image of student's answer.
        Your task is to evaluate the graph in the image based on the provided grading criteria and output a list indicating which criteria are fulfilled. You may also consider the text within the image when grading. [...] } \\ \\
        \texttt{    
            The output should be a list of binary values (1 or 0), where 1 indicates that the corresponding criterion is fulfilled and 0 means it is not fulfilled. For example, [...]
            } \\\\
        \texttt{ The output format must be a valid JSON.}\\
        \texttt{ Do not wrap the JSON output in markdown.}\\
        \texttt{ Do not include any explanatory text in the output.}\\\\
        
        \texttt{Task Description: Draw the initial situation for private tutoring at the secondary school level in Zurich [...] }\\\\
       
        \texttt{Grading Criteria: [start: demand curve less elastic than supply curve,
       consequence: supply curve shift to the left]}\\\\\\

        \textbf{User: }\texttt{Base64 graph} \\
        
        \textbf{Assistant: }\texttt{[0,0]} \\\\
    
        \textbf{User: }\texttt{Base64 graph} \\
        
        \textbf{Assistant: }\texttt{[0,1]} \\\\
    
        \textbf{User: }\texttt{Base64 graph} \\
        
        \textbf{Assistant: }\texttt{[1,1]} \\\\
    
        \textbf{User: }\texttt{Test base64 graph} 

         \\
         \hline
    \end{tabular}
    
\end{table}

\subsection{VLLMs}\label{section:result-vllms}
In addition to our specialized meta-learning models, we explored the use of VLLMs for automated grading. We evaluated multiple VLLMs, including \texttt{gpt-4.1}, \texttt{gpt-4.1-mini}, \texttt{gpt-4.1-nano}, \texttt{gpt-4o}, \texttt{gpt-4o-mini}, and \texttt{o4-mini}. All models were accessed via API using a low temperature setting of 0.1 to ensure more deterministic outputs (with slight variability in the outputs) except for \texttt{o4-mini} that doesn't accept temperature. 

A structured prompt (see Table~\ref{tab:llm-prompt}) was provided that included the task description, a set of grading criteria, and several example input-output pairs for the few-shot learning scenario. Unlike the meta-learning experiments, that relied on fixed N-way settings, VLLMs do not require a fixed number of classes. However, to ensure a fair comparison, we adopted the same data sampling strategy used in the meta-learning experiments. Specifically, we used the support examples to construct few-shot user/assistant pairs in the prompt while the query examples were used for evaluation.

Table~\ref{tab:updated-llm_results} summarizes the accuracy of various VLLMs under 2-way and 3-way classification tasks. Generally, more powerful model variants achieve higher accuracy, though this trend is not strictly consistent across all settings. For example, performance typically improves from \texttt{gpt-4.1-nano} and \texttt{gpt-4.1-mini} to \texttt{gpt-4.1}, with the exception of 3-way 2-shot and 4-shot scenarios. Similarly, \texttt{gpt-4o} mostly outperforms \texttt{gpt-4o-mini}, though an exception is observed at 2-way 4-shot, where \texttt{gpt-4o-mini} exceeds \texttt{gpt-4o}.

Among all models, \texttt{gpt-4.1} demonstrates consistently the strongest performance in 2-way tasks across all shot, achieving the highest overall accuracy of 53.09\% at 2-way, 4-shot. In the more complex 3-way setting, \texttt{gpt-4o} yields the best performance in the 1-shot and 2-shot settings, while \texttt{o4-mini} achieves the highest accuracy in the 3-way 4-shot configuration, with an accuracy of 50.00\%.

\begin{figure}[!h]
    \centering
    \includegraphics[width=0.8\columnwidth]{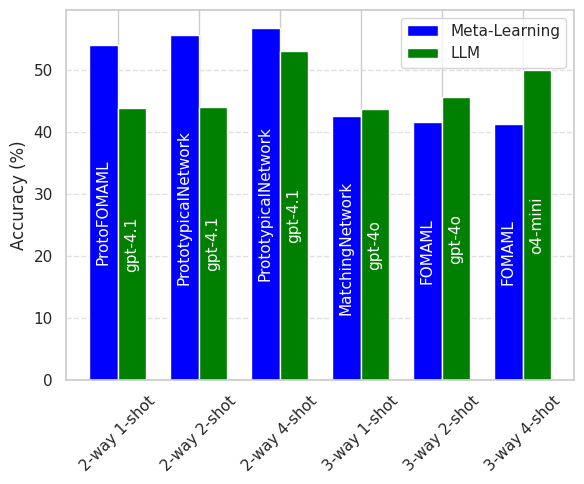}
    \caption{Comparison of best-performing meta-learning models and VLLMs across different N-way, K-shot classification settings.}
    \label{fig:meta-vs-llm-barplot}
\end{figure}

Although further prompt engineering and hyperparameter tuning
are required, these early results suggest that VLLMs have the
potential to achieve competitive and even higher performance
on the task of automated grading in some settings. A comparison of VLLMs and meta-learning models (Figure~\ref{fig:meta-vs-llm-barplot}) reveals that in 2-way tasks, the top-performing meta-learning models outperform the best-performing VLLMs across all shots. In contrast, in 3-way settings, the best-performing VLLMs outperform the meta-learning models.

We also examined the performance of VLLMs across individual module assignments in Figures~\ref{fig:results_llm_by_mas_nway2} and~\ref{fig:results_llm_by_mas_nway3}. For 2-way tasks (Figure~\ref{fig:results_llm_by_mas_nway2}), LLMs show often strong performance on tasks VWL7-HS21/22\_4, VWL8-FS22\_1, and VWL7-HS21/22\_8, while performance is lowest for VWL9-FS22\_2 and VWL8-FS22\_3. Similarly, in 3-way tasks (Figure~\ref{fig:results_llm_by_mas_nway3}), LLMs consistently perform well on VWL8-FS22\_17 and tend to struggle with VWL9-FS22\_2. These results suggest that, as with meta-learning models, certain tasks are inherently easier or harder for VLLMs to learn.

\section{Discussion} \label{section:discussion}

A major limitation of this study is that graph annotations were performed by a single teacher. This introduces potential for human error and subjective bias, which can influence model training and evaluation outcomes. A more robust and standard annotation approach would involve multiple expert annotators, ideally preceded by consensus on a shared grading rubric. Each annotator would independently label the graphs, and a majority vote or inter-annotator agreement metric would determine the final label. However, due to budget constraints and the limited availability of expert annotators in our institution, we were unable to implement this approach.

Another important limitation is the size and diversity of the dataset. Expanding the dataset by using additional courses from more recent years could improve model generalization and performance. This is part of our future research plan. Moreover, the dataset exhibits a strong class imbalance in different levels of graph correctness, which affects model performance. While we experimented with merging underrepresented classes to mitigate this, the adjustment did not significantly improve performance. This issue highlights the need for a more balanced and representative dataset in future.

\begin{figure}[!h]
    \centering
    \includegraphics[width=\columnwidth]{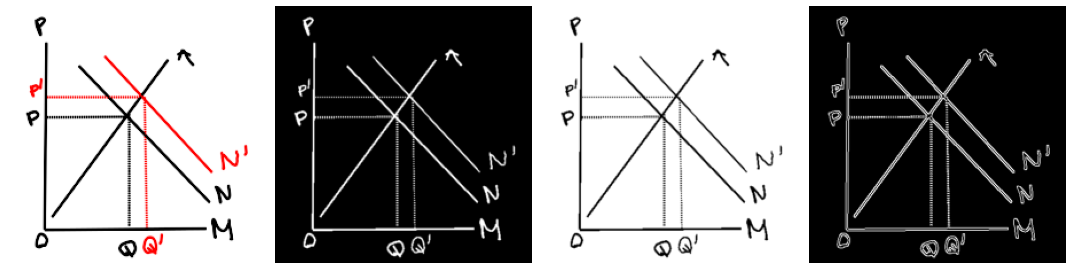}
    \caption{Comparison of image transformations applied to the original image. From left to right: the original image, thresholding, thresholding and inversion, and canny edge detection.}
    \label{fig:image_transformations}
\end{figure}

While our study used color images drawn on a digital canvas, real-world student submissions--often photographed with smartphones--can vary greatly in lighting, background, and quality, making color images less reliable in practice. However, our framework remains robust. By preprocessing the images and converting them into binary format, as shown in Fig. \ref{fig:image_transformations}, we can still train and evaluate our models effectively. We tested the framework on these image formats and found performance comparable to that obtained with the original color images.

Although VLLMs demonstrated competitive or even higher results compared to specially trained meta-learning algorithms, they still have limitations when used for autograding students' handwritten graph-based responses. In addition to their high costs, specially for image-based tasks, they also show instability when multiple test graphs are included in a single prompt--a technique intended to reduce token count and API calls. While the model performs well with a single test graph, reformulating the prompt to include multiple graphs in one call leads to inconsistent results (e.g., skipped or hallucinated responses). This highlights a limitation of VLLMs: despite clear prompts, the model struggles to accurately process and respond to multiple test graphs simultaneously.

The scope of VLLM experimentation was also limited. Due to resource constraints, we tested only a small number of prompt formats and hyperparameters. Future work should explore more diverse prompting strategies and models to optimize VLLM grading performance.

At the time of writing this manuscript, there is limited prior research on the automatic grading of students' handwritten graphs in math-based courses. This underscores the novelty of our approach but also emphasizes the need for further validation. While our study demonstrates the feasibility of using meta-learning and VLLMs for graph grading, future research should benchmark these methods against alternative approaches and assess their reliability across diverse, low-data scenarios.

Despite these limitations, our approach is adaptable. By expanding the dataset and involving multiple annotators, models can be retrained to improve accuracy and fairness. In the long term, this system could be integrated into the Moodle learning platform for relevant courses, assisting teachers by providing a second opinion on student grades or serving as an initial evaluation before manual assessment. Such an implementation would not only improve grading efficiency but also help instructors maintain consistent grading standards.

\section{Conclusion} \label{section:conclusion}

This study addressed the challenging and largely unexplored task of automatically grading students’ handwritten graphs in math-related courses. We developed multimodal meta-learning models that learn from few examples and compared their performance with VLLMs, including \texttt{gpt-4.1}, \texttt{gpt-4.1-mini}, \texttt{gpt-4.1-nano}, \texttt{gpt-4o}, \texttt{gpt-4o-mini}, and \texttt{4o-mini}. Using a real-world dataset from a distance university, we found that the best-performing meta-learning models (particularly ProtoFOMAML and Prototypical Network) outperform VLLMs in 2-way classification tasks, while in 3-way classification the top-performing VLLMs outperform the meat-learning models. Among VLLMs, \texttt{gpt-4.1} and \texttt{gpt-4o} often showed strong grading accuracy across few-shot scenarios.
However, despite their promising results, the practical use of VLLMs remains uncertain due to high computational costs, privacy and ethical considerations, and occasional inconsistencies. Further research is needed to address these challenges.

Future work will focus on expanding the dataset, involving multiple annotators for more reliable labeling, and more experiments with meta-learning models and VLLMs. 
Ultimately, integrating such systems into online learning platforms could enhance grading efficiency and consistency, offering educators a second opinion or an initial evaluation of students' graph-based responses.

\section{Acknowledgment}
This research was conducted with support from the Swiss National Science Foundation under the Spark grant (Grant No. 221412). 

\section{Competing Interests}
The authors have no relevant financial or non-financial interests to disclose.

\section{Author Contributions}
All authors contributed to the study conception and design. Material preparation, data collection and analysis were performed by Behnam Parsaeifard. The first draft of the manuscript was written by Behnam Parsaeifard and all authors commented on previous versions of the manuscript. All authors read and approved the final manuscript.


%





\ifCLASSOPTIONcaptionsoff
  \newpage
\fi
\vspace{0.05in}
\bibliographystyle{IEEEtran}
\bibliography{IEEEabrv,./main}

\begin{thebibliography}{10}
\providecommand{\url}[1]{#1}
\csname url@samestyle\endcsname
\providecommand{\newblock}{\relax}
\providecommand{\bibinfo}[2]{#2}
\providecommand{\BIBentrySTDinterwordspacing}{\spaceskip=0pt\relax}
\providecommand{\BIBentryALTinterwordstretchfactor}{4}
\providecommand{\BIBentryALTinterwordspacing}{\spaceskip=\fontdimen2\font plus
\BIBentryALTinterwordstretchfactor\fontdimen3\font minus \fontdimen4\font\relax}
\providecommand{\BIBforeignlanguage}[2]{{%
\expandafter\ifx\csname l@#1\endcsname\relax
\typeout{** WARNING: IEEEtran.bst: No hyphenation pattern has been}%
\typeout{** loaded for the language `#1'. Using the pattern for}%
\typeout{** the default language instead.}%
\else
\language=\csname l@#1\endcsname
\fi
#2}}
\providecommand{\BIBdecl}{\relax}
\BIBdecl

\bibitem{corbett1994knowledge}
A.~T. Corbett and J.~R. Anderson, ``Knowledge tracing: Modeling the acquisition of procedural knowledge,'' \emph{User modeling and user-adapted interaction}, vol.~4, pp. 253--278, 1994.

\bibitem{callear2001caa}
D.~H. Callear, J.~Jerrams-Smith, and V.~Soh, ``Caa of short non-mcq answers,'' 2001.

\bibitem{sakaguchi2015effective}
K.~Sakaguchi, M.~Heilman, and N.~Madnani, ``Effective feature integration for automated short answer scoring,'' in \emph{Proceedings of the 2015 conference of the North American Chapter of the association for computational linguistics: Human language technologies}, 2015, pp. 1049--1054.

\bibitem{magooda2016vector}
A.~E. Magooda, M.~Zahran, M.~Rashwan, H.~Raafat, and M.~Fayek, ``Vector based techniques for short answer grading,'' in \emph{The twenty-ninth international flairs conference}, 2016.

\bibitem{zhang2022automatic}
L.~Zhang, Y.~Huang, X.~Yang, S.~Yu, and F.~Zhuang, ``An automatic short-answer grading model for semi-open-ended questions,'' \emph{Interactive learning environments}, vol.~30, no.~1, pp. 177--190, 2022.

\bibitem{erickson2020automated}
J.~A. Erickson, A.~F. Botelho, S.~McAteer, A.~Varatharaj, and N.~T. Heffernan, ``The automated grading of student open responses in mathematics,'' in \emph{Proceedings of the tenth international conference on learning analytics \& knowledge}, 2020, pp. 615--624.

\bibitem{lan2015mathematical}
A.~S. Lan, D.~Vats, A.~E. Waters, and R.~G. Baraniuk, ``Mathematical language processing: Automatic grading and feedback for open response mathematical questions,'' in \emph{Proceedings of the second (2015) ACM conference on learning@ scale}, 2015, pp. 167--176.

\bibitem{mikolov2013efficient}
\BIBentryALTinterwordspacing
T.~Mikolov, K.~Chen, G.~S. Corrado, and J.~Dean, ``Efficient estimation of word representations in vector space,'' in \emph{International Conference on Learning Representations}, 2013. [Online]. Available: \url{https://api.semanticscholar.org/CorpusID:5959482}
\BIBentrySTDinterwordspacing

\bibitem{pennington2014glove}
J.~Pennington, R.~Socher, and C.~D. Manning, ``Glove: Global vectors for word representation,'' in \emph{Proceedings of the 2014 conference on empirical methods in natural language processing (EMNLP)}, 2014, pp. 1532--1543.

\bibitem{devlin2018bert}
J.~Devlin, M.-W. Chang, K.~Lee, and K.~Toutanova, ``Bert: Pre-training of deep bidirectional transformers for language understanding,'' \emph{arXiv preprint arXiv:1810.04805}, 2018.

\bibitem{reimers2019sentence}
N.~Reimers and I.~Gurevych, ``Sentence-bert: Sentence embeddings using siamese bert-networks,'' \emph{arXiv preprint arXiv:1908.10084}, 2019.

\bibitem{sung2019pre}
C.~Sung, T.~Dhamecha, S.~Saha, T.~Ma, V.~Reddy, and R.~Arora, ``Pre-training bert on domain resources for short answer grading,'' in \emph{Proceedings of the 2019 Conference on Empirical Methods in Natural Language Processing and the 9th International Joint Conference on Natural Language Processing (EMNLP-IJCNLP)}, 2019, pp. 6071--6075.

\bibitem{condor2020exploring}
A.~Condor, ``Exploring automatic short answer grading as a tool to assist in human rating,'' in \emph{Artificial Intelligence in Education: 21st International Conference, AIED 2020, Ifrane, Morocco, July 6--10, 2020, Proceedings, Part II 21}.\hskip 1em plus 0.5em minus 0.4em\relax Springer, 2020, pp. 74--79.

\bibitem{baral2021improving}
S.~Baral, A.~F. Botelho, J.~A. Erickson, P.~Benachamardi, and N.~T. Heffernan, ``Improving automated scoring of student open responses in mathematics.'' \emph{International Educational Data Mining Society}, 2021.

\bibitem{shen2021mathbert}
J.~T. Shen, M.~Yamashita, E.~Prihar, N.~Heffernan, X.~Wu, B.~Graff, and D.~Lee, ``Mathbert: A pre-trained language model for general nlp tasks in mathematics education,'' \emph{arXiv preprint arXiv:2106.07340}, 2021.

\bibitem{hamad2016detailed}
K.~Hamad and M.~Kaya, ``A detailed analysis of optical character recognition technology,'' \emph{International Journal of Applied Mathematics Electronics and Computers}, no. Special Issue-1, pp. 244--249, 2016.

\bibitem{zhang2017watch}
J.~Zhang, J.~Du, S.~Zhang, D.~Liu, Y.~Hu, J.~Hu, S.~Wei, and L.~Dai, ``Watch, attend and parse: An end-to-end neural network based approach to handwritten mathematical expression recognition,'' \emph{Pattern Recognition}, vol.~71, pp. 196--206, 2017.

\bibitem{zhang2018track}
J.~Zhang, J.~Du, and L.~Dai, ``Track, attend, and parse (tap): An end-to-end framework for online handwritten mathematical expression recognition,'' \emph{IEEE Transactions on Multimedia}, vol.~21, no.~1, pp. 221--233, 2018.

\bibitem{liu2024ai}
T.~Liu, J.~Chatain, L.~Kobel-Keller, G.~Kortemeyer, T.~Willwacher, and M.~Sachan, ``Ai-assisted automated short answer grading of handwritten university level mathematics exams,'' \emph{arXiv preprint arXiv:2408.11728}, 2024.

\bibitem{kortemeyer2023toward}
G.~Kortemeyer, ``Toward ai grading of student problem solutions in introductory physics: A feasibility study,'' \emph{Physical Review Physics Education Research}, vol.~19, no.~2, p. 020163, 2023.

\bibitem{baral2025drawedumath}
S.~Baral, L.~Lucy, R.~Knight, A.~Ng, L.~Soldaini, N.~T. Heffernan, and K.~Lo, ``Drawedumath: Evaluating vision language models with expert-annotated students' hand-drawn math images,'' \emph{arXiv preprint arXiv:2501.14877}, 2025.

\bibitem{baral2023auto}
S.~Baral, A.~Botelho, A.~Santhanam, A.~Gurung, L.~Cheng, and N.~Heffernan, ``Auto-scoring student responses with images in mathematics.''\hskip 1em plus 0.5em minus 0.4em\relax The Proceedings of the 16th International Conference on Educational Data Mining., 2023.

\bibitem{10.7717/peerj-cs.103}
\BIBentryALTinterwordspacing
A.~Meurer, C.~P. Smith, M.~Paprocki, O.~\v{C}ert\'{i}k, S.~B. Kirpichev, M.~Rocklin, A.~Kumar, S.~Ivanov, J.~K. Moore, S.~Singh, T.~Rathnayake, S.~Vig, B.~E. Granger, R.~P. Muller, F.~Bonazzi, H.~Gupta, S.~Vats, F.~Johansson, F.~Pedregosa, M.~J. Curry, A.~R. Terrel, v.~Rou\v{c}ka, A.~Saboo, I.~Fernando, S.~Kulal, R.~Cimrman, and A.~Scopatz, ``Sympy: symbolic computing in python,'' \emph{PeerJ Computer Science}, vol.~3, p. e103, Jan. 2017. [Online]. Available: \url{https://doi.org/10.7717/peerj-cs.103}
\BIBentrySTDinterwordspacing

\bibitem{zhang2022automaticmeta}
M.~Zhang, S.~Baral, N.~Heffernan, and A.~Lan, ``Automatic short math answer grading via in-context meta-learning,'' \emph{arXiv preprint arXiv:2205.15219}, 2022.

\bibitem{radford2021learning}
A.~Radford, J.~W. Kim, C.~Hallacy, A.~Ramesh, G.~Goh, S.~Agarwal, G.~Sastry, A.~Askell, P.~Mishkin, J.~Clark \emph{et~al.}, ``Learning transferable visual models from natural language supervision,'' in \emph{International conference on machine learning}.\hskip 1em plus 0.5em minus 0.4em\relax PmLR, 2021, pp. 8748--8763.

\bibitem{caraeni2024evaluating}
A.~Caraeni, A.~Scarlatos, and A.~Lan, ``Evaluating gpt-4 at grading handwritten solutions in math exams,'' \emph{arXiv preprint arXiv:2411.05231}, 2024.

\bibitem{heffernan2014assistments}
N.~T. Heffernan and C.~L. Heffernan, ``The assistments ecosystem: Building a platform that brings scientists and teachers together for minimally invasive research on human learning and teaching,'' \emph{International Journal of Artificial Intelligence in Education}, vol.~24, pp. 470--497, 2014.

\bibitem{smith2007overview}
R.~Smith, ``An overview of the tesseract ocr engine,'' in \emph{Ninth international conference on document analysis and recognition (ICDAR 2007)}, vol.~2.\hskip 1em plus 0.5em minus 0.4em\relax IEEE, 2007, pp. 629--633.

\bibitem{krizhevsky2012imagenet}
A.~Krizhevsky, I.~Sutskever, and G.~E. Hinton, ``Imagenet classification with deep convolutional neural networks,'' \emph{Advances in neural information processing systems}, vol.~25, 2012.

\bibitem{zeiler2014visualizing}
M.~D. Zeiler and R.~Fergus, ``Visualizing and understanding convolutional networks,'' in \emph{Computer Vision--ECCV 2014: 13th European Conference, Zurich, Switzerland, September 6-12, 2014, Proceedings, Part I 13}.\hskip 1em plus 0.5em minus 0.4em\relax Springer, 2014, pp. 818--833.

\bibitem{simonyan2015very}
K.~Simonyan and A.~Zisserman, ``Very deep convolutional networks for large-scale image recognition,'' in \emph{3rd International Conference on Learning Representations (ICLR 2015)}.\hskip 1em plus 0.5em minus 0.4em\relax Computational and Biological Learning Society, 2015.

\bibitem{szegedy2015going}
C.~Szegedy, W.~Liu, Y.~Jia, P.~Sermanet, S.~Reed, D.~Anguelov, D.~Erhan, V.~Vanhoucke, and A.~Rabinovich, ``Going deeper with convolutions,'' in \emph{Proceedings of the IEEE conference on computer vision and pattern recognition}, 2015, pp. 1--9.

\bibitem{he2016deep}
K.~He, X.~Zhang, S.~Ren, and J.~Sun, ``Deep residual learning for image recognition,'' in \emph{Proceedings of the IEEE conference on computer vision and pattern recognition}, 2016, pp. 770--778.

\bibitem{vinyals2016matching}
O.~Vinyals, C.~Blundell, T.~Lillicrap, D.~Wierstra \emph{et~al.}, ``Matching networks for one shot learning,'' \emph{Advances in neural information processing systems}, vol.~29, 2016.

\bibitem{snell2017prototypical}
J.~Snell, K.~Swersky, and R.~Zemel, ``Prototypical networks for few-shot learning,'' \emph{Advances in neural information processing systems}, vol.~30, 2017.

\bibitem{sung2018learning}
F.~Sung, Y.~Yang, L.~Zhang, T.~Xiang, P.~H. Torr, and T.~M. Hospedales, ``Learning to compare: Relation network for few-shot learning,'' in \emph{Proceedings of the IEEE conference on computer vision and pattern recognition}, 2018, pp. 1199--1208.

\bibitem{finn2017model}
C.~Finn, P.~Abbeel, and S.~Levine, ``Model-agnostic meta-learning for fast adaptation of deep networks,'' in \emph{International conference on machine learning}.\hskip 1em plus 0.5em minus 0.4em\relax PMLR, 2017, pp. 1126--1135.

\bibitem{Triantafillou2020Meta-Dataset:}
\BIBentryALTinterwordspacing
E.~Triantafillou, T.~Zhu, V.~Dumoulin, P.~Lamblin, U.~Evci, K.~Xu, R.~Goroshin, C.~Gelada, K.~Swersky, P.-A. Manzagol, and H.~Larochelle, ``Meta-dataset: A dataset of datasets for learning to learn from few examples,'' in \emph{International Conference on Learning Representations}, 2020. [Online]. Available: \url{https://openreview.net/forum?id=rkgAGAVKPr}
\BIBentrySTDinterwordspacing

\end{thebibliography}
\end{document}